\documentclass[10pt,twocolumn,letterpaper]{article}

\usepackage[pagenumbers]{cvpr} 

\usepackage{multirow}
\providecommand{\rc}[1]{#1}
\usepackage{colortbl}
\newcommand{\gcell}{\cellcolor[gray]{0.93}}
\usepackage{graphicx}

\definecolor{cvprblue}{rgb}{0.21,0.49,0.74}
\usepackage[pagebackref,breaklinks,colorlinks,allcolors=cvprblue]{hyperref}

\def\confName{CVPR}
\def\confYear{2027}

\title{CREDIT: Compositional Reliability Learning for Noisy Composed Image Retrieval}

\author{
Tiancai Zhou$^{1,*}$ \quad
Yongxiang Li$^{2,*}$ \quad
Suimei Liu$^{2}$ \quad
Zhong Yuan$^{3}$ \\
Dezhong Peng$^{1}$ \quad
Xi Peng$^{3}$ \quad
Xu Wang$^{1,\dagger}$ \\[3pt]
$^{1}$College of Computer Science, Sichuan University \\
$^{2}$School of Computing and Artificial Intelligence, Southwest Jiaotong University \\
$^{3}$School of Artificial Intelligence, Sichuan University \\[2pt]
}

\begin{document}
\maketitle
\begin{abstract}
Composed Image Retrieval (CIR) aims to retrieve a target image that satisfies user intent based on a multimodal query consisting of a reference image and a modification text. However, real-world training data are often corrupted by Noisy Triplet Correspondence (NTC), where the semantic correspondence among the reference image, modification text, and target image becomes unreliable. Existing methods implicitly assume that the correctness of triplet correspondence can be determined by fixed decision rules. However, under NTC, the authenticity of semantic correspondence is difficult to determine accurately, and different triplets exhibit varying degrees of semantic reliability rather than binary correctness. Consequently, deterministic confidence is insufficient to faithfully characterize correspondence reliability, often leading to overconfident supervision for unreliable triplets and progressively reinforcing noisy semantics during compositional representation learning. To address this challenge, we propose a \textbf{C}ompositional \textbf{R}eliability learning framework with \textbf{E}vidential uncertainty and \textbf{D}iffus\textbf{I}on-guided refinemen\textbf{T} (\textbf{CREDIT}), which explicitly quantifies the uncertainty of triplet correspondence to estimate its reliability and suppress the propagation of noisy supervision. CREDIT consists of two complementary components: The Evidential Compositional Uncertainty Learning (ECUL) module explicitly models correspondence uncertainty via evidential learning, enabling uncertainty-aware reliability estimation to suppress noisy supervision. Building upon the reliable supervision provided by ECUL, the Diffusion-enhanced Reference Refinement (DRR) module further refines reference representations through text-conditioned diffusion, producing more robust compositional query features. Extensive experiments demonstrate that CREDIT consistently outperforms existing methods under various settings.
\end{abstract}    
\section{Introduction}

\begin{figure*}[!t]
\centering
\includegraphics[
        width=\linewidth,
    ]{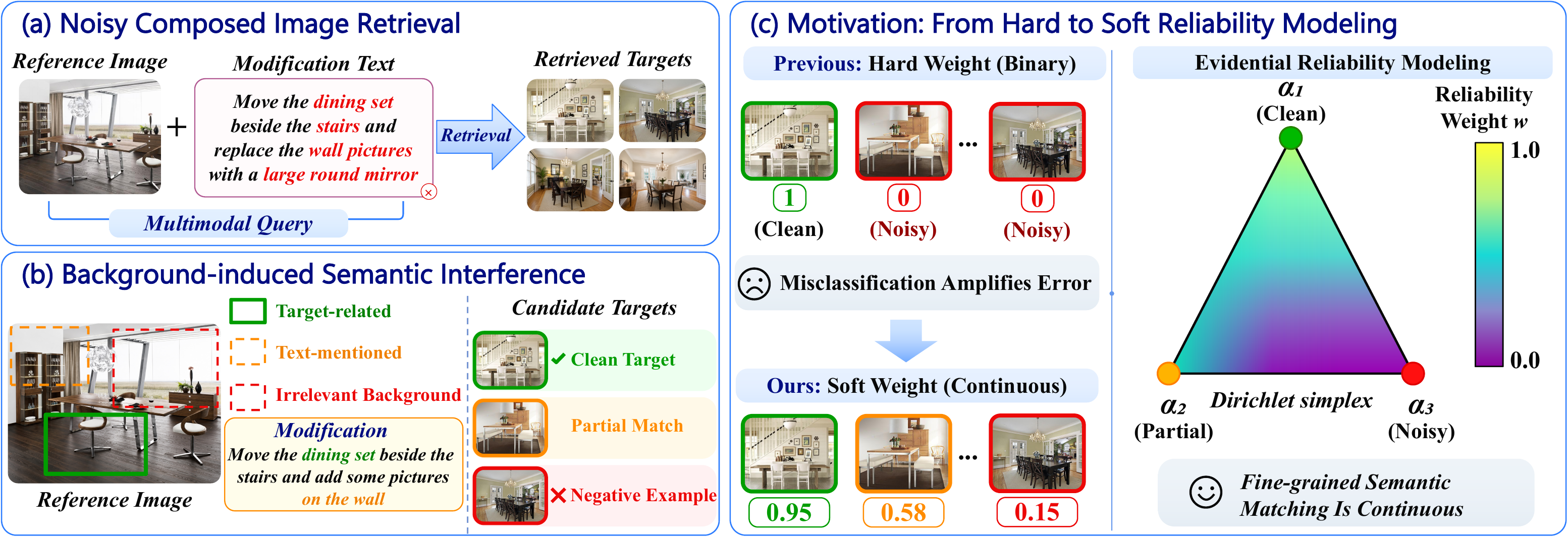}
\caption{
(a) Noisy Triplet Correspondence Paradigm in CIR.
(b) Background-induced semantic interference shows that irrelevant regions in the reference image may disturb reference-text composition.
(c) Motivation of Evidential Uncertainty Modeling.
}
\label{fig:motivation}
\end{figure*}

In recent years, with the rapid growth of multimedia data \cite{huang2025enhancing,10.1145/3240508.3240549,Tian2025OpenMR} and advances in multimodal learning techniques \cite{bi-etal-2025-llava,Liu2025SETransformerAH,bi2025prism, NEURIPS2025_1d6dfe3f}, Composed Image Retrieval (CIR) \cite{HUD,chen2025offset,11604175} has attracted increasing attention. Unlike conventional image retrieval, CIR enables flexible retrieval through multimodal queries consisting of a reference image and a modification text, where the text specifies the desired semantic modification of the reference image. The retrieval system aims to identify the target image that best satisfies the composed query. Owing to its capability of expressing fine-grained retrieval intent, CIR has been widely applied in information retrieval \cite{11017756,tian2025coremmrag,10.1145/3442381.3449986}, image editing \cite{lu2023tf,10.1145/3746027.3754546,lu2024robust}, and recommendation systems \cite{wu2021fashion,guo2018dialog,goenka2022fashionvlp}.

Despite remarkable progress, existing CIR methods \cite{8953387,li2025aaai-encoder} generally assume that the correspondence among the reference image, modification text, and target image is reliable during training. However, real-world data inevitably suffer from Noisy Triplet Correspondence (NTC), where the annotated correspondence among the three modalities becomes ambiguous or even incorrect, as illustrated in Figure~\ref{fig:motivation}(a). Such correspondence noise may originate from visually irrelevant background information in reference images or inaccurate textual descriptions that fail to faithfully characterize the target image. Consequently, 
different training triplets may exhibit substantially different degrees of semantic reliability, making it difficult to confidently infer correspondence correctness from observed matching signals alone.
Under this circumstance, accurately determining whether a triplet truly represents valid semantic correspondence becomes challenging. 

Existing robust methods \cite{TME,INTENT,HABIT} attempt to mitigate NTC by identifying and suppressing unreliable triplets, but typically rely on deterministic confidence estimation or hard decision rules to supervise compositional representation learning. Such strategies implicitly assume that triplet reliability can be inferred with high certainty from deterministic matching cues and separated using fixed decision boundaries. This assumption becomes fragile under NTC, where ambiguous triplets may exhibit similar matching confidence despite substantially different semantic reliability. Consequently, deterministic confidence alone cannot explicitly characterize how certain the model is about a correspondence relation, often leading to overconfident supervision for unreliable triplets and progressively reinforcing noisy semantics throughout optimization. 
Moreover, reliable correspondence estimation itself depends on the quality of multimodal composition. In practice, reference images often contain abundant background content unrelated to the modification text. As shown in Figure~\ref{fig:motivation}(b), irrelevant regions, such as distant outdoor scenery and decorative background elements, introduce semantic interference during reference-text interaction. Directly composing the entire visual representation with textual modifications inevitably propagates background noise into the composed query representation, making semantic evidence less discriminative and further increasing the difficulty of reliable correspondence estimation.
Therefore, robust learning under NTC requires not only estimating correspondence reliability with an explicit notion of uncertainty, but also reducing semantic interference before such reliability is assessed.

To address these challenges, we propose a \textbf{C}ompositional \textbf{R}eliability learning framework with \textbf{E}vidential uncertainty and \textbf{D}iffus\textbf{I}on-guided refinemen\textbf{T} (\textbf{CREDIT}) for noisy composed image retrieval. CREDIT consists of two complementary components: \emph{Evidential Compositional Uncertainty Learning (ECUL)} and \emph{Diffusion-enhanced Reference Refinement (DRR)}. ECUL first performs noise-aware reference-text composition through a Decoupled Residual Fusion Encoder (DRFE), which separates target-related visual information from irrelevant background regions via text-guided routing and adaptively preserves useful contextual cues. Building upon the resulting composed representation, ECUL further introduces Triplet Uncertainty Modeling (TUM), whose motivation is illustrated in Figure~\ref{fig:motivation}(c). Specifically, TUM formulates query-to-gallery matching evidence with a Dirichlet distribution under the Subjective Logic framework. Rather than relying solely on deterministic matching confidence,
TUM explicitly quantifies correspondence uncertainty and estimates the semantic reliability of each triplet according to the collected evidence, enabling reliable triplets to contribute stronger supervision while suppressing noisy correspondence during contrastive learning. Based on the reliable supervision established by ECUL, DRR further refines reference representations through a text-conditioned feature diffusion process that performs feature-level perturbation and denoising before multimodal composition, thereby producing more robust compositional query representations.

The main contributions of this paper are summarized as follows:
\begin{itemize}
\item We revisit NTC in composed image retrieval from the perspective of semantic correspondence reliability, showing that correspondence reliability cannot be reliably inferred from deterministic matching confidence alone due to the intrinsic uncertainty of semantic correspondence under noisy supervision.
\item We propose CREDIT, a compositional reliability learning framework that establishes reliable supervision by explicitly modeling correspondence uncertainty through evidential learning and further enhances compositional representation via diffusion-guided reference refinement, enabling robust multimodal composition under NTC.
\item Extensive experiments on CIR benchmarks demonstrate that CREDIT consistently achieves superior retrieval performance and exhibits strong robustness against NTC compared with existing state-of-the-art methods.
\end{itemize}
\section{Related Work}
\noindent\textbf{CIR with Noisy Triplet Correspondence.}
Current CIR approaches generally follow two paradigms: early works relying on separate feature extraction \cite{8953387, 10.1145/3404835.3462967} and recent advances leveraging pre-trained vision-language models \cite{pmlr-v139-radford21a} for joint learning \cite{10888153, 10890642, li2025finecirexplicitparsingfinegrained}. While these methods achieve superior results, they rely on the ideal assumption of well-aligned annotations. In practice, large-scale datasets inevitably contain NTC due to annotation errors or semantic ambiguity. While some recent methods \cite{Han_2023_CVPR, zhao2024mitigating, 10657023, 10653726, LI2025113304, 11710259} have explored the general noisy correspondence problem, they are primarily tailored for standard bi-modal pairs. Consequently, these strategies are ill-equipped to tackle the unique structural complexities posed by NTC in CIR. To address this, recent pioneering work \cite{TME, HABIT} has begun to tackle the NTC problem directly through sample selection and realignment strategies. However, these methods primarily focus on resolving cross-modal mismatches, while overlooking semantic interference during reference-text interaction.

\begin{figure*}[t]
    \centering
    \includegraphics[
        width=\linewidth,
    ]{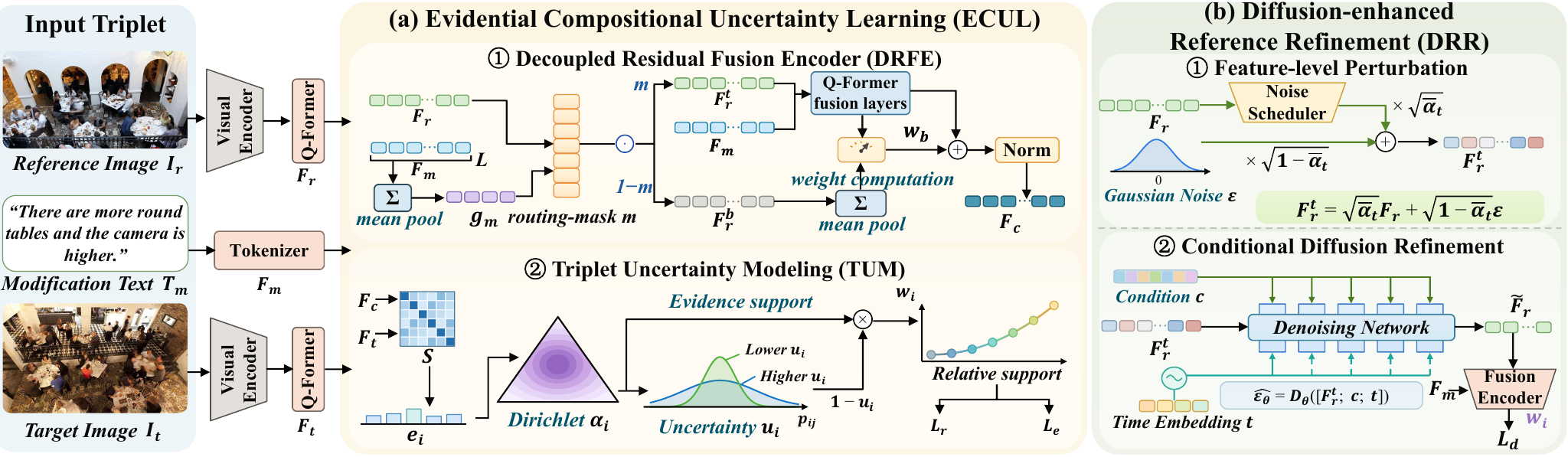}
    \caption{Overview of the CREDIT framework. The framework consists of two modules: (a) Evidential Compositional Uncertainty Learning (ECUL) and (b) Diffusion-enhanced Reference Refinement (DRR). ECUL integrates a Decoupled Residual Fusion Encoder and Triplet Uncertainty Modeling to reduce background interference during reference-text fusion and estimate triplet reliability under noisy supervision. DRR refines the reference representation through feature perturbation and denoising, and optimizes the refined composed query with a complementary contrastive loss to improve robustness under NTC.}
    \label{fig:framework}
\end{figure*}

\noindent\textbf{Uncertainty Learning.} Evidential Deep Learning (EDL) explicitly quantifies epistemic uncertainty by parameterizing categorical probabilities with a Dirichlet distribution. Early works \cite{sensoy2018evidential, amini2020deep} primarily developed conventional EDL frameworks for reliable uni-modal predictions. Recent advances leverage uncertainty-based filtering to improve model robustness against label noise and unreliable supervision \cite{han2021trusted, wang2021combating, bao2022evidential, 10888672}. While most research assumes noise exists solely in uni-modal labels, large-scale multimodal datasets frequently exhibit Noisy Correspondence (NC) due to annotation errors. Addressing such cross-modal mismatches is highly challenging, requiring the distinction between reliable and unreliable alignment signals during training \cite{huang2021learning, zhao2024mitigating, Han_2023_CVPR}. Recent work has begun to tackle this problem through cross-modal evidence extraction, improving the noise robustness of retrieval models. While some evidence-based methods \cite{qin2022deep} effectively mitigate bi-modal false positives, they mainly aim to improve standard image-text retrieval performance rather than addressing robustness under the NTC setting in CIR. Consequently, existing pair-wise strategies remain inadequate for tackling the unique challenges posed by noisy triplet structures and compositional ambiguity.
\section{Method}
\subsection{Problem Formulation}
Composed Image Retrieval (CIR) aims to retrieve the target image that matches a multimodal query composed of a reference image and a modification text. Given a training triplet set with NTC, we denote it as $\mathcal{D}=\{(I_r^i,T_m^i,I_t^i)\}_{i=1}^{N}$, where $I_r$, $T_m$, and $I_t$ represent the reference image, modification text, and target image. Ideally, $T_m$ describes the intended visual transformation from $I_r$ to $I_t$, and the retrieval model learns a function $\mathcal{R}_{\theta}$ that ranks the corresponding target image ahead of other gallery candidates. However, in practical CIR datasets, the elements within a triplet may be imperfectly aligned, leading to noisy triplets. In this paper, we consider two main noise sources: modality-inherent noise and cross-modal correspondence noise. Following TME~\cite{TME}, we simulate NTC by randomly selecting a subset of training triplets according to a noise ratio $\sigma$ and corrupting one component of each selected triplet. Under this setting, our goal is to learn a robust retrieval function $\mathcal{R}_{\theta}$ that captures reliable compositional semantics while suppressing noisy triplet supervision.

\subsection{Evidential Compositional Uncertainty Learning}

\textbf{Decoupled Residual Fusion Encoder.}
As illustrated in Figure~\ref{fig:framework}(a), the Decoupled Residual Fusion Encoder (DRFE) is designed to perform noise-aware composition between the reference image and the modification text. Since the reference image may contain irrelevant background regions, directly fusing all visual queries with the text can introduce modality-inherent noise. To alleviate this issue, DRFE decouples the reference visual queries into text-relevant visual components and text-irrelevant background components, and then adaptively recombines useful contextual information.

Given a reference image $I_r$, a modification text $T_m$, and a target image $I_t$, we first obtain the reference visual queries and target image features through Q-Former:
\begin{equation}
    \mathbf{F}_r
    =
    \operatorname{Q\text{-}Former}
    \left(
    \mathbf{Q},
    \Phi_I(I_r)
    \right),
    \mathbf{F}_t
    =
    \operatorname{Q\text{-}Former}
    \left(
    \mathbf{Q},
    \Phi_I(I_t)
    \right),
\end{equation}
where $\Phi_I(\cdot)$ denotes the frozen image encoder, $\mathbf{Q}\in\mathbb{R}^{N\times D}$ is a set of learnable query tokens, and $\mathbf{F}_r,\mathbf{F}_t\in\mathbb{R}^{N\times D}$, $N$ is the number of query tokens and $D$ is the feature dimension.

Meanwhile, the modification text is encoded as $\mathbf{F}_m=\Phi_T(T_m),$ where $\Phi_T(\cdot)$ denotes the text embedding encoder and $\mathbf{F}_m\in\mathbb{R}^{L\times D}$, where $L$ is the text sequence length. To obtain the global modification semantics, we average the textual features along the token dimension:
\begin{equation}
    \boldsymbol{g}_m
    =
    \frac{1}{L}
    \sum_{l=1}^{L}
    \boldsymbol{f}_{m,l},
\end{equation}
where $\boldsymbol{f}_{m,l}$ denotes the textual feature in $\mathbf{F}_m$.

To estimate the relevance between each visual query and the modification text, we compute a text-guided routing mask:
\begin{equation}
    \boldsymbol{m}
    =
    \sigma
    \left(
    \cos(\mathbf{F}_r,\boldsymbol{g}_m)
    {
    /\tau_r
    }
    \right),
\end{equation}
where $\tau_r$ is a temperature coefficient, $\sigma(\cdot)$ denotes the sigmoid function, and $\boldsymbol{m}\in\mathbb{R}^{N\times 1}$. Based on this mask, the reference visual queries are decoupled into a text-relevant visual component and a text-irrelevant background component:
\begin{equation}
    \mathbf{F}_r^{t}
    =
    \mathbf{F}_r\odot\boldsymbol{m},
    \quad
    \mathbf{F}_r^{b}
    =
    \mathbf{F}_r\odot(1-\boldsymbol{m}),
\end{equation}
where $\odot$ denotes element-wise multiplication.

Then, only the text-relevant visual queries $\mathbf{F}_r^{t}$ and the textual features $\mathbf{F}_m$ are fed into the fusion layers of Q-Former:
\begin{equation}
    \mathbf{F}_{q}
    =
    \operatorname{Q\text{-}Former}
    (\mathbf{F}_r^{t},\mathbf{F}_m),
\end{equation}
where $\mathbf{F}_{q}\in\mathbb{R}^{N\times D}$ denotes the target-aware composed representation.

Although the background component may contain irrelevant information, it can still provide useful context for retrieval. Therefore, we generate a dynamic residual weight according to the similarity between the composed representation and the pooled background feature:
\begin{equation}
    w_b
    =
    1-
    \sigma
    \left(
    \cos
    \left(
    \mathbf{F}_{q},
    \operatorname{pool}(\mathbf{F}_r^{b})
    \right){
    / \tau_r
    }
    \right),
\end{equation}
where $\operatorname{pool}(\cdot)$ denotes mean pooling. The composed query feature is obtained by residual recombination:
\begin{equation}
    \mathbf{F}_{c}
    =
    \operatorname{Norm}
    \left(
    \mathbf{F}_{q}
    +
    w_b\cdot
    \operatorname{pool}(\mathbf{F}_r^{b})
    \right).
\end{equation}
The resulting feature $\mathbf{F}_{c}$ is used for subsequent triplet uncertainty modeling.

\noindent\textbf{Triplet Uncertainty Modeling.} 
Although DRFE suppresses modality-inherent noise during semantic interaction, NTC may still make the query-target supervision unreliable. A deterministic similarity score can only indicate the relative matching strength, but cannot explicitly reflect how confident the model is about this matching relation. To address this issue, we introduce an Triplet Uncertainty Modeling (TUM), as shown in Figure~\ref{fig:framework}(a), which models the matching relation between a composed query and target images from the perspective of evidential deep learning.

\textbf{\textit{Triplet-level Evidence Construction.}}
Given a mini-batch containing $B$ triplets, we first compute the similarity matrix $\boldsymbol{S}\in\mathbb{R}^{B\times B}$ between the composed query features and target image features. To reduce evidence inflation caused by uniformly elevated matching responses, we center the transformed matching responses and map them into non-negative evidence:
\begin{equation}
\left\{
\begin{aligned}
    z_{ij}
    &= \tanh(S_{ij}),
    \qquad
    \bar z_i = \frac{1}{B}\sum_{k=1}^{B}z_{ik}, \\
    e_{ij}
    &= \exp\!\left(
    \frac{z_{ij}-\bar z_i}{\tau_e}
    \right),
\end{aligned}
\right.
\end{equation}
where $z_{ij}$ denotes the transformed matching response,
$\bar z_i$ is its mean over the $B$ candidate targets,
and $\tau_e$ is the evidence temperature. The evidence vector $\boldsymbol{e}_i=[e_{i1},e_{i2},\cdots,e_{iB}]$ describes the support assigned to all candidate targets for the $i$-th composed query.

\textbf{\textit{Reliability Estimation for NTC.}}
Following Subjective Logic~\cite{josang2016subjective}, the collected evidence parameterizes a Dirichlet distribution over candidate targets, with $\alpha_{ij}=e_{ij}+1$ and distribution strength $A_i=\sum_{j=1}^{B}\alpha_{ij}$. The belief mass and overall uncertainty are computed as: 
\begin{equation}
    b_{ij}
    =
    \frac{e_{ij}}{A_i}
    =
    \frac{\alpha_{ij}-1}{A_i},
    \qquad
    u_i=\frac{B}{A_i},
\end{equation}
where $u_i+\sum_{j=1}^{B}b_{ij}=1$. Based on the evidence support for the observed target, we define a continuous reliability weight:
\begin{equation}
    w_i
    =
    \left(
    (1-u_i)
    \frac{e_{ii}}{\sum_{j=1}^{B}e_{ij}}
    \right)^2.
\end{equation}
This formulation assigns higher reliability weights $w_i$ to clean triplets whose target images faithfully reflect the fine-grained semantic modifications specified by the text, while reducing the contributions of ambiguous or noisy triple correspondences.

To concentrate evidential supervision on well-supported correspondences, we compare the observed target with its strongest competing target and construct a trusted mask:
\begin{equation}
\left\{
\begin{aligned}
    \Delta_i
    &=
    \frac{
    e_{ii}
    -
    \max_{j \neq i}e_{ij}
    }{e_{ii}}, \\[4pt]
    \chi_i
    &=
    \mathbb{I}\!\left[
    (1-u_i)\Delta_i>\delta
    \right].
\end{aligned}
\right.
\end{equation}
where $\delta$ is the evidence acceptance threshold. 

\textbf{\textit{Evidential Objective.}}
Let $y_{ij}=\mathbb{I}[j=i]$ encode the observed query-target assignment. We formulate the evidential objective as the weighted expected squared matching error under the predicted Dirichlet distribution:
\begin{equation}
\begin{aligned}
    \mathcal{L}_e
    &=
    \frac{1}{\sum_{i=1}^{B}w_i}
    \sum_{i=1}^{B}w_i\chi_i
    \int
    \|\boldsymbol{y}_i-\boldsymbol{p}_i\|_2^2
    D(\boldsymbol{p}_i\mid\boldsymbol{\alpha}_i)
    \,d\boldsymbol{p}_i \\
    &\hspace{-1.5em}=
    \frac{1}{\sum_{i=1}^{B}w_i}
    \sum_{i=1}^{B}w_i\chi_i
    \sum_{j=1}^{B}
    \Bigg[
    \left(
    y_{ij}-\frac{\alpha_{ij}}{A_i}
    \right)^2
    +
    \frac{
    \alpha_{ij}(A_i-\alpha_{ij})
    }{
    A_i^2(A_i+1)
    }
    \Bigg].
\end{aligned}
\end{equation}
The trusted mask concentrates evidential supervision on well-supported correspondences.

\begin{table*}[!t]
\centering
\caption{Performance of ordinary methods and robust methods (gray shading) on the FashionIQ validation set. The best and second-best results are marked in bold and underlined, respectively.}
\label{tab:fashioniq}
{\small
\renewcommand{\arraystretch}{0.80}
\begin{tabular}{c l cc cc cc cc c}
\toprule[1.25pt]
\multirow{2}{*}{Noise} & \multirow{2}{*}{Methods}
& \multicolumn{2}{c}{Dress}
& \multicolumn{2}{c}{Shirt}
& \multicolumn{2}{c}{Toptee}
& \multicolumn{2}{c}{Average}
& \multirow{2}{*}{AVG.} \\
\cmidrule(lr){3-4}
\cmidrule(lr){5-6}
\cmidrule(lr){7-8}
\cmidrule(lr){9-10}
& & R@10 & R@50 & R@10 & R@50 & R@10 & R@50 & R@10 & R@50 & \\
\midrule
\multirow{8}{*}{0\%}
& CALA \cite{10.1145/3626772.3657823} (SIGIR'24) & 42.38 & 66.08 & 46.76 & 68.16 & 50.93 & 73.42 & 46.69 & 69.22 & 57.96 \\
& SPRC \cite{bai2024sentence} (ICLR'24) & 49.18 & 72.43 & 55.64 & 73.89 & {59.35} & 78.58 & 54.72 & 74.97 & 64.85 \\
& \rc{RCL} \cite{10050111} (TPAMI'23) & \rc{48.79} & \rc{\underline{72.68}} & \rc{55.89} & \rc{73.90} & \rc{56.91} & \rc{77.41} & \rc{53.86} & \rc{74.66} & \rc{64.26} \\
& \rc{RDE} \cite{qin2024noisy} (CVPR'24) & \rc{47.84} & \rc{71.89} & \rc{54.37} & \rc{73.55} & \rc{56.91} & \rc{77.21} & \rc{53.04} & \rc{74.22} & \rc{63.63} \\
& \gcell \rc{TME} \cite{TME} (CVPR'25) & \gcell \rc{49.73} & \gcell \rc{71.69} & \gcell \rc{{56.43}} & \gcell \rc{74.44} & \gcell \rc{59.31} & \gcell \rc{\underline{78.94}} & \gcell \rc{55.15} & \gcell \rc{75.02} & \gcell \rc{65.09} \\
& \gcell \rc{HABIT} \cite{HABIT} (AAAI'26) & \gcell \rc{{49.99}} & \gcell \rc{{72.38}} & \gcell \rc{\underline{56.62}} & \gcell \rc{{74.68}} & \gcell \rc{\underline{59.51}} & \gcell \rc{{78.53}} & \gcell \rc{\underline{55.38}} & \gcell \rc{\underline{75.20}} & \gcell \rc{\underline{65.29}} \\
& \gcell \rc{INTENT} \cite{INTENT} (AAAI'26) & \gcell \rc{\underline{50.32}} & \gcell \rc{72.10} & \gcell \rc{56.32} & \gcell \rc{\underline{74.93}} & \gcell \rc{59.28} & \gcell \rc{78.45} & \gcell \rc{{55.31}} & \gcell \rc{{75.16}} & \gcell \rc{{65.24}} \\
& \gcell \rc{\textbf{CREDIT (Ours)}} & \gcell \rc{\textbf{51.51}} & \gcell \rc{\textbf{74.17}} & \gcell \rc{\textbf{56.92}} & \gcell \rc{\textbf{76.20}} & \gcell \rc{\textbf{60.48}} & \gcell \rc{\textbf{80.83}} & \gcell \rc{\textbf{56.30}} & \gcell \rc{\textbf{77.07}} & \gcell \rc{\textbf{66.69}} \\
\midrule
\multirow{8}{*}{20\%}
& CALA \cite{10.1145/3626772.3657823} (SIGIR'24) & 29.05 & 51.36 & 35.28 & 56.23 & 36.05 & 58.24 & 33.46 & 55.28 & 44.37 \\
& SPRC \cite{bai2024sentence} (ICLR'24) & 39.81 & 62.22 & 48.58 & 66.29 & 50.48 & 70.58 & 46.29 & 66.36 & 56.33 \\
& \rc{RCL} \cite{10050111} (TPAMI'23) & \rc{47.05} & \rc{70.65} & \rc{53.14} & \rc{71.74} & \rc{55.28} & \rc{75.62} & \rc{51.82} & \rc{72.67} & \rc{62.25} \\
& \rc{RDE} \cite{qin2024noisy} (CVPR'24) & \rc{44.62} & \rc{68.91} & \rc{50.74} & \rc{69.09} & \rc{52.12} & \rc{73.38} & \rc{49.16} & \rc{70.46} & \rc{59.81} \\
& \gcell \rc{TME} \cite{TME} (CVPR'25) & \gcell \rc{49.03} & \gcell \rc{70.35} & \gcell \rc{\underline{55.84}} & \gcell \rc{73.16} & \gcell \rc{57.22} & \gcell \rc{78.23} & \gcell \rc{54.03} & \gcell \rc{73.91} & \gcell \rc{63.97} \\
& \gcell \rc{HABIT} \cite{HABIT} (AAAI'26) & \gcell \rc{\underline{49.63}} & \gcell \rc{{71.34}} & \gcell \rc{{55.67}} & \gcell \rc{{73.19}} & \gcell \rc{\underline{58.14}} & \gcell \rc{{78.32}} & \gcell \rc{\underline{54.48}} & \gcell \rc{{74.28}} & \gcell \rc{\underline{64.38}} \\
& \gcell \rc{INTENT} \cite{INTENT} (AAAI'26) & \gcell \rc{{49.32}} & \gcell \rc{\underline{71.43}} & \gcell \rc{55.32} & \gcell \rc{\underline{73.57}} & \gcell \rc{{58.01}} & \gcell \rc{\underline{78.46}} & \gcell \rc{{54.22}} & \gcell \rc{\underline{74.49}} & \gcell \rc{{64.36}} \\
& \gcell \rc{\textbf{CREDIT (Ours)}} & \gcell \rc{\textbf{50.67}} & \gcell \rc{\textbf{72.29}} & \gcell \rc{\textbf{56.08}} & \gcell \rc{\textbf{74.93}} & \gcell \rc{\textbf{59.20}} & \gcell \rc{\textbf{79.96}} & \gcell \rc{\textbf{55.32}} & \gcell \rc{\textbf{75.73}} & \gcell \rc{\textbf{65.52}} \\
\midrule
\multirow{8}{*}{50\%}
& CALA \cite{10.1145/3626772.3657823} (SIGIR'24) & 20.77 & 40.95 & 26.69 & 46.57 & 27.03 & 46.81 & 24.83 & 44.78 & 34.80 \\
& SPRC \cite{bai2024sentence} (ICLR'24) & 35.94 & 57.16 & 42.25 & 61.63 & 44.98 & 64.76 & 41.06 & 61.19 & 51.12 \\
& \rc{RCL} \cite{10050111} (TPAMI'23) & \rc{43.68} & \rc{66.44} & \rc{50.74} & \rc{69.19} & \rc{52.63} & \rc{73.84} & \rc{49.01} & \rc{69.82} & \rc{59.42} \\
& \rc{RDE} \cite{qin2024noisy} (CVPR'24) & \rc{41.30} & \rc{64.75} & \rc{47.06} & \rc{66.34} & \rc{50.13} & \rc{70.63} & \rc{46.16} & \rc{67.24} & \rc{56.70} \\
& \gcell \rc{TME} \cite{TME} (CVPR'25) & \gcell \rc{46.26} & \gcell \rc{68.27} & \gcell \rc{{53.09}} & \gcell \rc{71.88} & \gcell \rc{55.07} & \gcell \rc{{76.59}} & \gcell \rc{51.47} & \gcell \rc{72.25} & \gcell \rc{61.86} \\
& \gcell \rc{HABIT} \cite{HABIT} (AAAI'26) & \gcell \rc{{47.33}} & \gcell \rc{{69.71}} & \gcell \rc{\underline{53.72}} & \gcell \rc{\underline{72.55}} & \gcell \rc{{56.51}} & \gcell \rc{\underline{77.00}} & \gcell \rc{\underline{52.52}} & \gcell \rc{{73.09}} & \gcell \rc{{62.80}} \\
& \gcell \rc{INTENT} \cite{INTENT} (AAAI'26) & \gcell \rc{\underline{47.99}} & \gcell \rc{\textbf{71.24}} & \gcell \rc{52.78} & \gcell \rc{{72.48}} & \gcell \rc{\underline{56.79}} & \gcell \rc{76.23} & \gcell \rc{\underline{52.52}} & \gcell \rc{\underline{73.32}} & \gcell \rc{\underline{62.92}} \\
& \gcell \rc{\textbf{CREDIT (Ours)}} & \gcell \rc{\textbf{48.14}} & \gcell \rc{\underline{70.25}} & \gcell \rc{\textbf{53.93}} & \gcell \rc{\textbf{73.65}} & \gcell \rc{\textbf{57.32}} & \gcell \rc{\textbf{78.12}} & \gcell \rc{\textbf{53.13}} & \gcell \rc{\textbf{74.01}} & \gcell \rc{\textbf{63.57}} \\
\midrule
\multirow{8}{*}{80\%}
& CALA \cite{10.1145/3626772.3657823} (SIGIR'24) & 14.28 & 30.59 & 19.73 & 35.82 & 19.48 & 36.10 & 17.83 & 34.17 & 26.00 \\
& SPRC \cite{bai2024sentence} (ICLR'24) & 28.41 & 50.77 & 36.21 & 54.37 & 35.90 & 56.96 & 33.51 & 54.03 & 43.77 \\
& \rc{RCL} \cite{10050111} (TPAMI'23) & \rc{38.82} & \rc{60.54} & \rc{45.44} & \rc{64.38} & \rc{47.42} & \rc{68.38} & \rc{43.89} & \rc{64.43} & \rc{54.16} \\
& \rc{RDE} \cite{qin2024noisy} (CVPR'24) & \rc{37.63} & \rc{59.63} & \rc{43.62} & \rc{62.12} & \rc{46.10} & \rc{66.50} & \rc{42.45} & \rc{62.75} & \rc{52.60} \\
& \gcell \rc{TME} \cite{TME} (CVPR'25) & \gcell \rc{41.45} & \gcell \rc{64.35} & \gcell \rc{47.30} & \gcell \rc{68.20} & \gcell \rc{51.25} & \gcell \rc{73.23} & \gcell \rc{46.67} & \gcell \rc{68.60} & \gcell \rc{57.63} \\
& \gcell \rc{HABIT} \cite{HABIT} (AAAI'26) & \gcell \rc{{42.04}} & \gcell \rc{{65.20}} & \gcell \rc{{50.12}} & \gcell \rc{\underline{69.77}} & \gcell \rc{{52.92}} & \gcell \rc{{73.61}} & \gcell \rc{{48.36}} & \gcell \rc{{69.53}} & \gcell \rc{{58.94}} \\
& \gcell \rc{INTENT} \cite{INTENT} (AAAI'26) & \gcell \rc{\underline{42.07}} & \gcell \rc{\underline{65.58}} & \gcell \rc{\underline{50.38}} & \gcell \rc{{69.41}} & \gcell \rc{\underline{53.09}} & \gcell \rc{\underline{73.91}} & \gcell \rc{\underline{48.51}} & \gcell \rc{\underline{69.63}} & \gcell \rc{\underline{59.07}} \\
& \gcell \rc{\textbf{CREDIT (Ours)}} & \gcell \rc{\textbf{44.47}} & \gcell \rc{\textbf{66.83}} & \gcell \rc{\textbf{51.08}} & \gcell \rc{\textbf{71.88}} & \gcell \rc{\textbf{53.85}} & \gcell \rc{\textbf{74.81}} & \gcell \rc{\textbf{49.80}} & \gcell \rc{\textbf{71.17}} & \gcell \rc{\textbf{60.49}} \\
\bottomrule[1.25pt]
\end{tabular}
}
\end{table*}

\begin{table*}[!t]
\centering
\caption{Performance of ordinary methods and robust methods (gray shading) on the CIRR test set. The best and second-best results are marked in bold and underlined, respectively.}
\label{tab:cirr}
{\small
\renewcommand{\arraystretch}{0.80}
\begin{tabular}{c l cccc ccc c}
\toprule[1.25pt]
\multirow{2}{*}{Noise} & \multirow{2}{*}{Methods}
& \multicolumn{4}{c}{R@K}
& \multicolumn{3}{c}{Rsub@K}
& \multirow{2}{*}{Avg(R@5, Rsub@1)} \\
\cmidrule(lr){3-6}
\cmidrule(lr){7-9}
& & K=1 & K=5 & K=10 & K=50 & K=1 & K=2 & K=3 & \\
\midrule
\multirow{8}{*}{0\%}
& CALA \cite{10.1145/3626772.3657823} (SIGIR'24) & 49.11 & 81.21 & 89.59 & 98.00 & 76.27 & 91.04 & 96.46 & 78.74 \\
& SPRC \cite{bai2024sentence} (ICLR'24) & 51.96 & 82.12 & 89.74 & 97.69 & {80.65} & 92.31 & 96.60 & 81.39 \\
& \rc{RCL} \cite{10050111} (TPAMI'23) & \rc{53.16} & \rc{82.41} & \rc{90.12} & \rc{\underline{98.34}} & \rc{79.57} & \rc{92.02} & \rc{96.87} & \rc{80.99} \\
& \rc{RDE} \cite{qin2024noisy} (CVPR'24) & \rc{51.81} & \rc{82.02} & \rc{90.60} & \rc{97.93} & \rc{78.17} & \rc{91.90} & \rc{96.70} & \rc{80.10} \\
& \gcell \rc{TME} \cite{TME} (CVPR'25) & \gcell \rc{\underline{53.42}} & \gcell \rc{82.99} & \gcell \rc{90.24} & \gcell \rc{98.15} & \gcell \rc{\textbf{81.04}} & \gcell \rc{\underline{92.58}} & \gcell \rc{\underline{96.94}} & \gcell \rc{\textbf{82.02}} \\
& \gcell \rc{HABIT} \cite{HABIT} (AAAI'26) & \gcell \rc{52.71} & \gcell \rc{{82.64}} & \gcell \rc{{90.63}} & \gcell \rc{98.19} & \gcell \rc{\underline{80.99}} & \gcell \rc{\textbf{92.77}} & \gcell \rc{\textbf{97.00}} & \gcell \rc{\underline{81.82}} \\
& \gcell \rc{INTENT} \cite{INTENT} (AAAI'26) & \gcell \rc{53.37} & \gcell \rc{\underline{83.16}} & \gcell \rc{\underline{90.73}} & \gcell \rc{98.22} & \gcell \rc{80.24} & \gcell \rc{{92.37}} & \gcell \rc{{96.89}} & \gcell \rc{{81.70}} \\
& \gcell \rc{\textbf{CREDIT (Ours)}} & \gcell \rc{\textbf{54.08}} & \gcell \rc{\textbf{83.54}} & \gcell \rc{\textbf{91.30}} & \gcell \rc{\textbf{98.51}} & \gcell \rc{79.16} & \gcell \rc{92.05} & \gcell \rc{96.82} & \gcell \rc{81.35} \\
\midrule
\multirow{8}{*}{20\%}
& CALA \cite{10.1145/3626772.3657823} (SIGIR'24) & 41.33 & 72.70 & 82.84 & 94.34 & 71.66 & 88.15 & 94.94 & 72.18 \\
& SPRC \cite{bai2024sentence} (ICLR'24) & 45.90 & 75.86 & 83.52 & 93.37 & 78.10 & {91.40} & 96.05 & 76.98 \\
& \rc{RCL} \cite{10050111} (TPAMI'23) & \rc{50.43} & \rc{81.11} & \rc{88.82} & \rc{96.68} & \rc{77.52} & \rc{90.80} & \rc{95.71} & \rc{79.31} \\
& \rc{RDE} \cite{qin2024noisy} (CVPR'24) & \rc{49.23} & \rc{78.63} & \rc{86.80} & \rc{95.78} & \rc{76.58} & \rc{90.31} & \rc{96.07} & \rc{77.60} \\
& \gcell \rc{TME} \cite{TME} (CVPR'25) & \gcell \rc{{51.35}} & \gcell \rc{81.01} & \gcell \rc{88.53} & \gcell \rc{97.81} & \gcell \rc{\underline{78.46}} & \gcell \rc{91.25} & \gcell \rc{96.39} & \gcell \rc{\underline{79.74}} \\
& \gcell \rc{HABIT} \cite{HABIT} (AAAI'26) & \gcell \rc{\underline{51.68}} & \gcell \rc{{81.02}} & \gcell \rc{{89.24}} & \gcell \rc{97.81} & \gcell \rc{78.20} & \gcell \rc{\underline{91.66}} & \gcell \rc{\textbf{96.75}} & \gcell \rc{{79.61}} \\
& \gcell \rc{INTENT} \cite{INTENT} (AAAI'26) & \gcell \rc{51.25} & \gcell \rc{\underline{81.36}} & \gcell \rc{\underline{90.02}} & \gcell \rc{\underline{98.05}} & \gcell \rc{77.95} & \gcell \rc{{91.40}} & \gcell \rc{\underline{96.46}} & \gcell \rc{79.66} \\
& \gcell \rc{\textbf{CREDIT (Ours)}} & \gcell \rc{\textbf{52.22}} & \gcell \rc{\textbf{82.51}} & \gcell \rc{\textbf{90.22}} & \gcell \rc{\textbf{98.36}} & \gcell \rc{\textbf{78.53}} & \gcell \rc{\textbf{92.02}} & \gcell \rc{\textbf{96.75}} & \gcell \rc{\textbf{80.52}} \\
\midrule
\multirow{8}{*}{50\%}
& CALA \cite{10.1145/3626772.3657823} (SIGIR'24) & 36.10 & 66.12 & 77.76 & 92.10 & 68.12 & 85.66 & 93.59 & 67.12 \\
& SPRC \cite{bai2024sentence} (ICLR'24) & 39.93 & 66.00 & 73.59 & 86.48 & 75.81 & 89.21 & 95.37 & 70.90 \\
& \rc{RCL} \cite{10050111} (TPAMI'23) & \rc{48.58} & \rc{77.45} & \rc{85.93} & \rc{94.70} & \rc{75.60} & \rc{89.28} & \rc{94.80} & \rc{76.52} \\
& \rc{RDE} \cite{qin2024noisy} (CVPR'24) & \rc{45.98} & \rc{75.30} & \rc{83.73} & \rc{94.48} & \rc{73.98} & \rc{88.99} & \rc{95.13} & \rc{74.64} \\
& \gcell \rc{TME} \cite{TME} (CVPR'25) & \gcell \rc{48.48} & \gcell \rc{78.94} & \gcell \rc{87.28} & \gcell \rc{96.99} & \gcell \rc{76.48} & \gcell \rc{90.07} & \gcell \rc{95.83} & \gcell \rc{77.71} \\
& \gcell \rc{HABIT} \cite{HABIT} (AAAI'26) & \gcell \rc{\underline{50.32}} & \gcell \rc{{79.63}} & \gcell \rc{{88.34}} & \gcell \rc{97.06} & \gcell \rc{76.84} & \gcell \rc{\underline{90.60}} & \gcell \rc{\underline{96.27}} & \gcell \rc{\underline{78.87}} \\
& \gcell \rc{INTENT} \cite{INTENT} (AAAI'26) & \gcell \rc{{49.78}} & \gcell \rc{\underline{79.64}} & \gcell \rc{\underline{88.99}} & \gcell \rc{\underline{97.37}} & \gcell \rc{\underline{77.18}} & \gcell \rc{{90.41}} & \gcell \rc{{96.00}} & \gcell \rc{{78.41}} \\
& \gcell \rc{\textbf{CREDIT (Ours)}} & \gcell \rc{\textbf{51.23}} & \gcell \rc{\textbf{80.89}} & \gcell \rc{\textbf{89.74}} & \gcell \rc{\textbf{97.88}} & \gcell \rc{\textbf{78.72}} & \gcell \rc{\textbf{91.52}} & \gcell \rc{\textbf{96.41}} & \gcell \rc{\textbf{79.81}} \\
\midrule
\multirow{8}{*}{80\%}
& CALA \cite{10.1145/3626772.3657823} (SIGIR'24) & 31.52 & 61.49 & 72.60 & 89.86 & 64.34 & 83.52 & 92.60 & 62.92 \\
& SPRC \cite{bai2024sentence} (ICLR'24) & 29.95 & 51.25 & 58.51 & 73.86 & 70.22 & 86.05 & 93.21 & 60.74 \\
& \rc{RCL} \cite{10050111} (TPAMI'23) & \rc{44.94} & \rc{74.43} & \rc{82.99} & \rc{92.31} & \rc{71.93} & \rc{86.84} & \rc{92.96} & \rc{73.18} \\
& \rc{RDE} \cite{qin2024noisy} (CVPR'24) & \rc{42.92} & \rc{71.30} & \rc{80.51} & \rc{92.96} & \rc{69.64} & \rc{85.86} & \rc{93.54} & \rc{70.47} \\
& \gcell \rc{TME} \cite{TME} (CVPR'25) & \gcell \rc{46.31} & \gcell \rc{75.78} & \gcell \rc{84.89} & \gcell \rc{95.83} & \gcell \rc{73.37} & \gcell \rc{88.02} & \gcell \rc{94.89} & \gcell \rc{74.58} \\
& \gcell \rc{HABIT} \cite{HABIT} (AAAI'26) & \gcell \rc{\underline{47.93}} & \gcell \rc{{76.84}} & \gcell \rc{{85.95}} & \gcell \rc{95.90} & \gcell \rc{\underline{74.87}} & \gcell \rc{{89.08}} & \gcell \rc{{95.21}} & \gcell \rc{{75.86}} \\
& \gcell \rc{INTENT} \cite{INTENT} (AAAI'26) & \gcell \rc{{47.90}} & \gcell \rc{\underline{78.13}} & \gcell \rc{\underline{87.04}} & \gcell \rc{\underline{96.47}} & \gcell \rc{{73.81}} & \gcell \rc{\underline{89.18}} & \gcell \rc{\underline{95.54}} & \gcell \rc{\underline{75.97}} \\
& \gcell \rc{\textbf{CREDIT (Ours)}} & \gcell \rc{\textbf{49.30}} & \gcell \rc{\textbf{79.37}} & \gcell \rc{\textbf{88.22}} & \gcell \rc{\textbf{97.33}} & \gcell \rc{\textbf{77.42}} & \gcell \rc{\textbf{90.53}} & \gcell \rc{\textbf{95.90}} & \gcell \rc{\textbf{78.40}} \\
\bottomrule[1.25pt]
\end{tabular}
}
\end{table*}

Following complementary contrastive learning~\cite{10050111}, we optimize the composed representation using the continuous supervision weights. For a mini-batch with $B$ triplets, the robust complementary contrastive loss is defined as:
\begin{equation}
\begin{aligned}
    \mathcal{L}_r
    =
    &-\frac{1}{\sum_{i=1}^{B}w_i}
    \sum_{i, j\ne i}^{B}w_i
    \log\left(
    1-
    \frac{
    \exp(S_{ij}/\tau)
    }{
    \sum_{k=1}^{B}\exp(S_{ik}/\tau)
    }
    \right),
\end{aligned}
\end{equation}
where $\tau$ is a learnable temperature parameter. All triplets participate in retrieval learning, with their contributions continuously regulated by the matching evidence and uncertainty.

\subsection{Diffusion-enhanced Reference Refinement}
Although DRFE alleviates background interference during reference-text composition, reliable correspondence estimation still requires the reference representation to remain stable under feature perturbations. To improve such robustness, we introduce a \emph{Diffusion-enhanced Reference Refinement (DRR)} module, as shown in Figure~\ref{fig:framework}(b). Diffusion is well suited to this purpose, as its forward-reverse formulation naturally couples controllable feature perturbation with text-conditioned semantic recovery. Specifically, DRR perturbs the reference feature at varying noise levels and learns to recover modification-relevant semantics conditioned on the modification text before multimodal composition, thereby yielding a more robust reference representation for subsequent retrieval. Let $\mathbf{F}_{r}\in\mathbb{R}^{N\times D}$ denote the reference image feature extracted by the image encoder. Given the modification text $T_m$, we first obtain a text condition $\boldsymbol{c}$ and expand it to the token dimension to guide the feature denoising process.

\noindent\textbf{Feature-level Perturbation.}
Inspired by feature denoising diffusion \cite{NEURIPS2020_4c5bcfec}, we perform the forward diffusion process in the reference feature space. During training, a diffusion timestep $t$ is randomly sampled from a small timestep range to avoid over-corrupting the high-level reference feature. The noisy reference feature is obtained as:
\begin{equation}
    \mathbf{F}_{r}^{t}
    =
    \sqrt{\bar{\alpha}_{t}}\mathbf{F}_{r}
    +
    \sqrt{1-\bar{\alpha}_{t}}\boldsymbol{\epsilon},
    \quad
    \boldsymbol{\epsilon}\sim\mathcal{N}(0,\mathbf{I}),
\end{equation}
where $\bar{\alpha}_{t}$ is the cumulative noise schedule coefficient. Following the standard diffusion formulation, it is defined as $\bar{\alpha}_{t}:=\prod_{s=1}^{t}\alpha_s=\prod_{s=1}^{t}\left(1-\beta_s\right)$,where $\beta_s$ denotes the noise variance schedule. In our implementation, the Gaussian noise is further scaled according to the feature distribution of $\mathbf{F}_{r}$ to avoid excessively disturbing the reference feature space.

\noindent\textbf{Conditional Diffusion Refinement.}
A lightweight conditional denoising network predicts the noise component from the noisy reference feature, the text condition, and the timestep embedding:
\begin{equation}
    \hat{\boldsymbol{\epsilon}}_{\theta}
    =
    D_{\theta}
    \left(
    [\mathbf{F}_{r}^{t};\boldsymbol{c};\boldsymbol{t}]
    \right),
\end{equation}
where $D_{\theta}(\cdot)$ denotes the feature noise predictor, and $\boldsymbol{t}$ is the timestep embedding. According to the predicted noise, the enhanced reference feature is estimated by:
\begin{equation}
    \widetilde{\mathbf{F}}_{r}
    =
    \frac{
    \mathbf{F}_{r}^{t}
    -
    \sqrt{1-\bar{\alpha}_{t}}
    \hat{\boldsymbol{\epsilon}}_{\theta}
    }{
    \sqrt{\bar{\alpha}_{t}}
    }.
\end{equation}

The enhanced reference feature $\widetilde{\mathbf{F}}_{r}$ is further fused with the modification text to obtain the diffusion-enhanced composed feature:
\begin{equation}
    \mathbf{F}_{d}
    =
    \Phi
    \left(
    \widetilde{\mathbf{F}}_{r},
    T_m
    \right),
\end{equation}
where $\Phi(\cdot)$ denotes the decoupled residual fusion encoder, and $\mathbf{F}_{d}$ denotes the composed feature generated from the diffusion-enhanced reference feature.

For the $i$-th diffusion-enhanced composed feature, its similarity to the $j$-th target image is denoted as $S_{ij}^{d}$. The diffusion-enhanced complementary contrastive loss is formulated as:
\begin{equation}
    \mathcal{L}_{d}
    =
    -\frac{1}{\sum_{i=1}^{B}w_i}
    \sum_{i, j\ne i}^{B}w_i
    \log
    \left(
    1-
    \frac{
    \exp(S_{ij}^{d}/\tau)
    }{
    \sum_{k}^{B}\exp(S_{ik}^{d}/\tau)
    }
    \right).
\end{equation}

\subsection{Training Objective}
Finally, the overall training objective of CREDIT is formulated as:
\begin{equation}
    \Theta^{*}
    =
    \arg\min_{\Theta}
    \left(
    \mathcal{L}_{e}
    +
    \mu \mathcal{L}_{r}
    +
    \kappa \mathcal{L}_{d}
    \right),
\end{equation}
where $\Theta$ denotes the trainable parameters of CREDIT, and $\Theta^{*}$ represents the optimized parameters. $\mu$ and $\kappa$ are trade-off hyperparameters that balance the contributions of the corresponding loss terms.

\begin{table}[t]
\centering
\caption{Ablation studies on FashionIQ and CIRR under 20\% NTC.}
\label{tab:ablation}
{\small
\renewcommand{\arraystretch}{0.80}
\setlength{\tabcolsep}{2.2pt}
\resizebox{0.98\columnwidth}{!}{%
\begin{tabular}{c l cc cc}
\toprule[1.25pt]
\multirow{2}{*}{No.}
& \multirow{2}{*}{Variants}
& \multicolumn{2}{c}{FashionIQ-Avg.}
& \multicolumn{2}{c}{CIRR-Avg.} \\
\cmidrule(lr){3-4}
\cmidrule(lr){5-6}
& 
& R@10 & R@50
& R@K & $\mathrm{R_{sub}@K}$ \\
\midrule

\multicolumn{6}{c}{
\textbf{\textit{(a) Ablation Study of Loss Functions}}
} \\
\midrule

1
& $\mathcal{L}_{r}$
& 53.39 & 74.31
& 80.09 & 88.67 \\

2
& $\mathcal{L}_{r}+\mathcal{L}_{e}$
& 54.24 & 75.19
& 80.48 & 88.73 \\

\midrule

\multicolumn{6}{c}{
\textbf{\textit{(b) Ablation Study of Model Modules}}
} \\
\midrule

3
& w/o DRFE
& 54.62 & 75.27
& 80.43 & 88.54 \\

4
& w/o TUM
& 53.31 & 74.47
& 80.23 & 88.36 \\

5
& w/o DRR
& 54.19 & 74.76
& 80.64 & 88.64 \\

\midrule

\multicolumn{2}{c}{\textbf{CREDIT (Ours)}}
& \textbf{55.32}
& \textbf{75.73}
& \textbf{80.83}
& \textbf{89.10} \\

\bottomrule[1.25pt]
\end{tabular}%
}
}
\end{table}
\section{Experiments}
\subsection{Experimental Settings}
\noindent\textbf{Datasets.} Following the common CIR protocol, we evaluate on FashionIQ~\cite{wu2021fashion} for fashion-domain retrieval and CIRR~\cite{Liu_2021_ICCV} for open-domain retrieval.

\noindent\textbf{Implementation Details.} We train CREDIT via AdamW with a learning rate of $1e‑5$ on a 32GB Tesla V100 GPU. Following~\cite{bai2024sentence}, we initialize the image and fusion encoders with BLIP-2~\cite{li2023blip2} pre-trained weights and adopt EVA-CLIP ViT-g/14~\cite{Fang2022EVAET} as the visual backbone. All input images are resized to $224 \times 224$ with a padding ratio of 1.25, and hyperparameters $\mu=1$, $\kappa=1$, $\tau_r=0.06$, $\tau_e=0.1$ are shared across both datasets. We adopt Recall@K (R@K) as the evaluation metric to judge whether the ground-truth target exists in top-K retrieved results: CIRR reports overall R@$\{1,5,10,50\}$ and subset R@$\{1,2,3\}$, while FashionIQ evaluates R@10 and R@50 over three categories (Dress, Shirt, Toptee).

\subsection{Comparison with State-of-the-Arts}
We compare CREDIT with representative CIR methods, including standard CIR models and noise-correspondence robust variants, on FashionIQ and CIRR under noise ratios of $0.0$, $0.2$, $0.5$, and $0.8$. The results are summarized in Table~\ref{tab:fashioniq} and Table~\ref{tab:cirr}. On FashionIQ, CREDIT achieves the best AVG. score across all noise settings, outperforming the strongest existing method by 1.40, 1.14, 0.65, and 1.42 percentage points, respectively. These consistent improvements demonstrate the robustness of CREDIT against NTC. On CIRR, CREDIT obtains the best $\mathrm{Avg}(\mathrm{R@5},\mathrm{Rsub@1})$ under noise ratios of $0.2$, $0.5$, and $0.8$, surpassing the strongest baselines by 0.78, 0.94, and 2.43 percentage points, respectively. Under the clean setting on CIRR, evidential weighting may down-weight hard but clean triplets with ambiguous matching evidence. This can weaken supervision for fine-grained discrimination within subsets.

\subsection{Ablation Study}

To verify the effectiveness of each component in CREDIT, we conduct ablation studies on FashionIQ and CIRR under 20\% NTC. The summarized results are reported in Table~\ref{tab:ablation}.

\textbf{Loss function ablation.} As shown in Table~\ref{tab:ablation}(a), introducing $\mathcal{L}_{e}$ improves performance over using $\mathcal{L}_{r}$ alone. This objective concentrates evidential supervision on well-supported correspondences, limiting the influence of unreliable assignments. Further incorporating $\mathcal{L}_{d}$ yields the best results on both datasets, improving FashionIQ R@10 from 53.39 to 55.32 and R@50 from 74.31 to 75.73, while CIRR R@K increases from 80.09 to 80.83. $\mathcal{L}_{d}$ complements evidential supervision by reinforcing modification-relevant semantics in diffusion-enhanced queries.

\textbf{Module ablation.} As shown in Table~\ref{tab:ablation}(b), removing any component degrades retrieval performance. Without DRFE, text-irrelevant background content can interfere with reference-text fusion. Rremoving TUM causes the largest overall drop, reducing FashionIQ R@10 from 55.32 to 53.31 and CIRR Rsub@K from 89.10 to 88.36.
Without its reliability weighting, unreliable triplets can receive excessive supervision and distort the learned modification semantics.
Removing DRR also decreases FashionIQ R@10 by 1.13 points, supporting the complementary role of reference refinement.

\begin{figure}[t]
    \centering
    \includegraphics[width=0.55\linewidth]
    {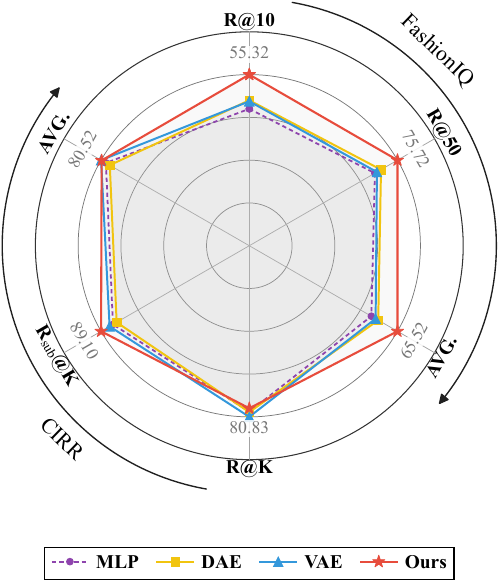}
    \caption{
        Comparison of feature enhancement methods under 20\% NTC.
        AVG. denotes the mean of R@10 and R@50 on FashionIQ,
        and of R@5 and Rsub@1 on CIRR.
    }
    \label{fig:feature_enhancement}
\end{figure}

\textbf{Feature enhancement ablation.}
We replace the diffusion-based refinement in DRR with a text-conditioned residual MLP, DAE \cite{10.1145/1390156.1390294}, or VAE \cite{2969442.2969628}. As shown in Figure~\ref{fig:feature_enhancement}, DRR achieves the best performance on FashionIQ, exceeding the strongest alternative, DAE, by 1.04 percentage points in AVG. On CIRR, DRR obtains the highest Rsub@K of 89.10 and an AVG. of 80.52, outperforming MLP and DAE by 0.23 and 0.44 points, respectively. Direct residual mapping lacks explicit denoising constraints, while autoencoding favor dominant visual content at the expense of subtle modification cues. By learning text-conditioned denoising across different noise levels, DRR encourages the recovery of modification-relevant semantics from perturbed features. These results support its effectiveness in refining compositional representations under NTC.

\subsection{Parameter Analysis}
\begin{figure}[!htbp]
\centering

\includegraphics[width=0.49\linewidth]
{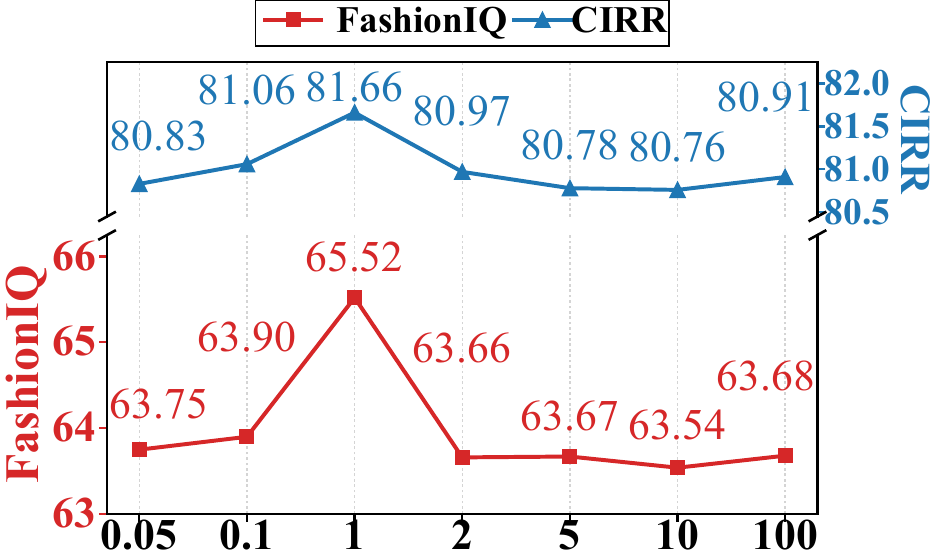}
\hfill
\includegraphics[width=0.49\linewidth]
{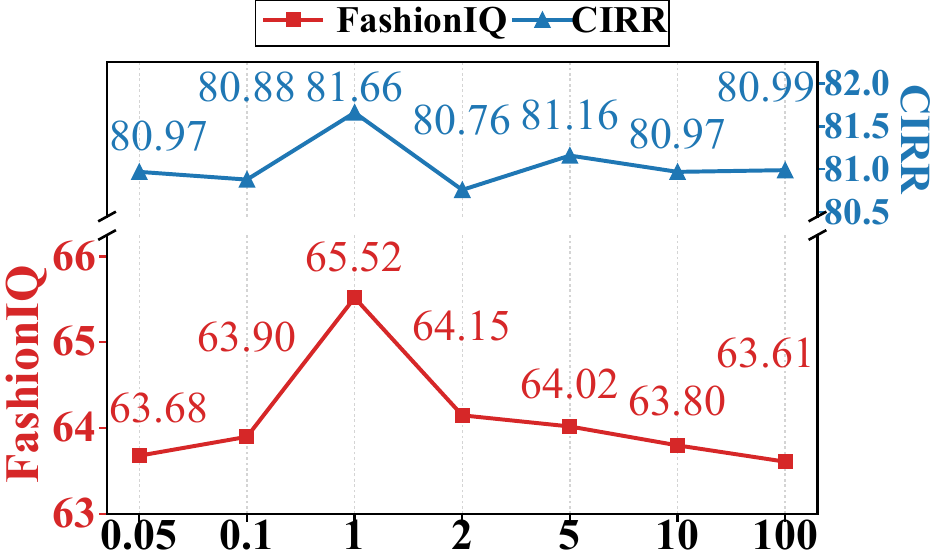}
\par

{\small
\makebox[0.49\linewidth][c]{(a) $\mu$}
\hfill
\makebox[0.49\linewidth][c]{(b) $\kappa$}
}

\caption{Parameter sensitivity analysis of CREDIT on the FashionIQ and CIRR validation sets with noise ratio $\sigma=0.2$.}
\label{fig:parametric_analysis}
\end{figure}

To assess CREDIT's sensitivity to the loss weights $\mu$ and $\kappa$, we analyze its performance on FashionIQ and CIRR validation sets, as shown in Figure~\ref{fig:parametric_analysis}. In Figure~\ref{fig:parametric_analysis}(a), increasing $\mu$, which weights the robust contrastive loss $\mathcal{L}_{r}$, from 0.05 to 1 improves performance on both datasets, while larger values lead to a slight decline. This suggests that moderate reliability-aware contrastive supervision helps suppress noisy correspondences, whereas excessive weighting may amplify residual supervision errors arising from imperfect reliability estimates. In Figure~\ref{fig:parametric_analysis}(b), $\kappa$, which controls the diffusion-enhanced complementary contrastive loss $\mathcal{L}_{d}$, shows a similar trend: performance first improves and then slightly drops. This indicates that diffusion-guided reference refinement benefits noisy composed image retrieval, but an overly large diffusion constraint may disturb reference-text semantic alignment. Therefore, we set $\mu=\kappa=1$ in our final model.

\subsection{Case Study}
\begin{figure}[!htbp]
    \centering
    \includegraphics[
        width=\linewidth
    ]{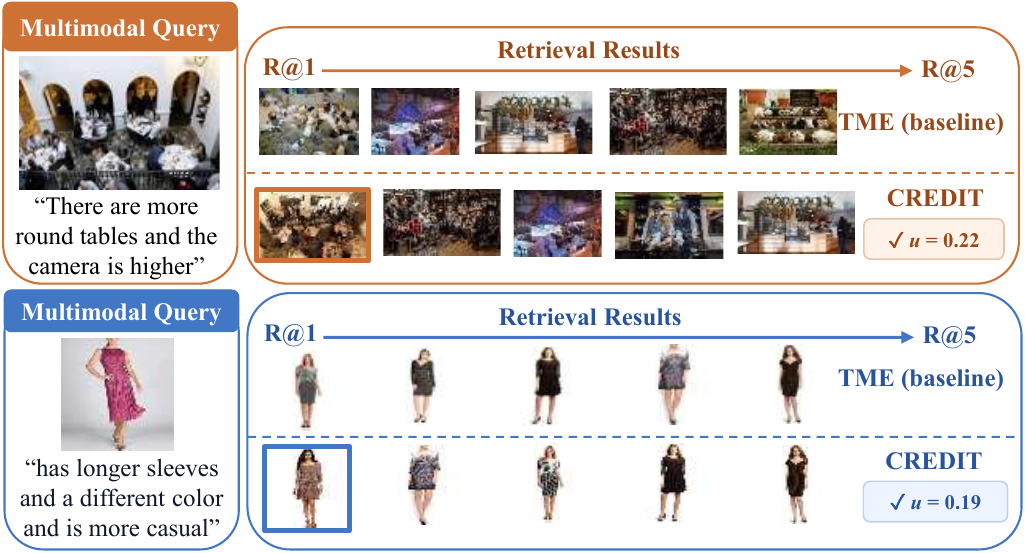}
    \caption{Case study comparison on CIRR and FashionIQ.}
    \label{fig:case_study}
\end{figure}

Figure~\ref{fig:case_study} compares the top-5 retrieval results of CREDIT and TME~\cite{TME} on CIRR and FashionIQ. On CIRR, CREDIT identifies the scene with more round tables and a higher camera viewpoint. On FashionIQ, it retrieves the dress with longer sleeves, a different color, and a more casual style. CREDIT ranks the correct target first in both cases, whereas TME fails to retrieve it within the displayed top-5 results. The corresponding uncertainty scores are $u=0.22$ and $u=0.19$, respectively. These examples illustrate CREDIT's ability to preserve relevant reference semantics while capturing the attribute and spatial changes specified by the modification text.

\section{Conclusion}
In this paper, we investigated NTC in composed image retrieval from the perspective of semantic correspondence reliability. We proposed CREDIT, a compositional reliability learning framework that contains two key modules. The ECUL module modeled triplet uncertainty via evidential learning and provided reliable supervision for robust contrastive learning. Meanwhile, the DRR module refined reference representations to enhance compositional robustness. Extensive experiments on FashionIQ and CIRR validated the effectiveness of CREDIT under NTC. In future work, we plan to investigate more adaptive uncertainty estimation mechanisms and extend our framework to more challenging real-world multimodal retrieval settings.

\section*{LLM Usage Statement}
We used OpenAI’s ChatGPT to refine the wording of author-written text, assist with portions of the experimental and analysis code, and help create figures based on our results. The authors reviewed all AI-assisted work and take full responsibility for the final text, claims, code, and figures.
{
    \small
    \bibliographystyle{ieeenat_fullname}
    \bibliography{main}
}


\end{document}